\documentclass[letterpaper, 10 pt, conference]{ieeeconf}  % Comment this line out if you need a4paper
\usepackage{xcolor}
\usepackage{booktabs}

\IEEEoverridecommandlockouts                              % This command is only needed if 
\title{\LARGE \bf
Vision-Based Fuzzy Control System for Smart Walkers: Enhancing Usability for Stroke Survivors with Unilateral Upper Limb Impairments
}

\author{\resizebox{\textwidth}{!}{Mahdi Chalaki$^{1,4}$, Amir Zakerimanesh$^{1}$, Abed Soleymani$^{1}$, Vivian Mushahwar$^{2,4}$, and Mahdi Tavakoli$^{1,3,4}$}%\textit{Senior Member, IEEE}% <-this % stops a space
\thanks{*This research was supported by the Canada Foundation for Innovation (CFI), the Natural Sciences and Engineering Research Council (NSERC) of Canada, the Canadian Institutes of Health Research (CIHR), Alberta Innovates, and the Government of Alberta’s grant to Centre for Autonomous Systems in Strengthening Future Communities}% <-this % stops a space
\thanks{$^{1}$M. Chalaki, A. Zakerimanesh, A. Soleymani, and M. Tavakoli are with the
Department of Electrical and Computer Engineering, University of
Alberta, Edmonton, AB, Canada. {\tt\small mahdichalaki, amir.zakerimanesh, zsoleymani, mahdi.tavakoli}@ualberta.ca }%
\thanks{$^{2}$V. Mushahwar is with the Department of Medicine, Division of Physical Medicine and Rehabilitation, University of Alberta
        {\tt\small vivian.mushahwar@ualberta.ca}}%
\thanks{$^{3}$M. Tavakoli is with the Department of Biomedical Engineering, University of Alberta}%
\thanks{$^{4}$M. Chalaki, M. Tavakoli, and V. Mushahwar are with the Institute for Smart Augmentative and Restorative Technologies and Health Innovations (iSMART), University of Alberta}%
}

\usepackage{booktabs}
\usepackage{multirow}
\usepackage{array}
\usepackage{graphicx}
\usepackage{cite}
\usepackage{subcaption}
\usepackage{url}
\usepackage{hyperref}

\begin{document}

\maketitle
\thispagestyle{empty}
\pagestyle{empty}

%%%%%%%%%%%%%%%%%%%%%%%%%%%%%%%%%%%%%%%%%%%%%%%%%%%%%%%%%%%%%%%%%%%%%%%%%%%%%%%%
\begin{abstract}

Mobility impairments, particularly those caused by stroke-induced hemiparesis, significantly impact independence and quality of life. 
Current smart walker controllers operate by using input forces from the user to control linear motion and input torques to dictate rotational movement; however, because they predominantly rely on user-applied torque exerted on the device handle as an indicator of user intent to turn, they fail to adequately accommodate users with unilateral upper limb impairments. This leads to increased physical strain and cognitive load. This paper introduces a novel smart walker equipped with a fuzzy control algorithm that leverages shoulder abduction angles to intuitively interpret user intentions using just one functional hand. By integrating a force sensor and stereo camera, the system enhances walker responsiveness and usability.
Experimental evaluations with five participants showed that the fuzzy controller outperformed the traditional admittance controller, reducing wrist torque while using the right hand to operate the walker by 12.65\% for left turns, 80.36\% for straight paths, and 81.16\% for right turns. Additionally, average user comfort ratings on a Likert scale increased from 1 to 4. Results confirmed a strong correlation between shoulder abduction angles and directional intent, with users reporting decreased effort and enhanced ease of use. This study contributes to assistive robotics by providing an adaptable control mechanism for smart walkers, suggesting a pathway towards enhancing mobility and independence for individuals with mobility impairments.

\noindent Project page: \textcolor{red}{\href{https://tbs-ualberta.github.io/fuzzy-sw/}{https://tbs-ualberta.github.io/fuzzy-sw/}} 

\end{abstract}

%%%%%%%%%%%%%%%%%%%%%%%%%%%%%%%%%%%%%%%%%%%%%%%%%%%%%%%%%%%%%%%%%%%%%%%%%%%%%%%%
\section{INTRODUCTION}

Bipedal locomotion is fundamental to our independence, quality of life, and mental well-being. It supports our self-esteem, social interactions, and ability to navigate various environments \cite{chalaki2024evaluating}. However, mobility tends to decline with age, primarily due to the deterioration of neurological, muscular, and skeletal systems \cite{paulo_isr-aiwalker_2017}. Yet, aging is not the sole cause of reduced mobility; conditions such as cardiovascular events (like strokes), spinal cord injuries, and diseases such as multiple sclerosis and Parkinson's, which are not always age-related, can also significantly impair mobility \cite{paulo_isr-aiwalker_2017}.

Among the most common causes of walking disabilities, stroke stands out as a leading factor. It is particularly severe, with as many as $88$\% of stroke survivors experiencing hemiparesis—a condition that results in muscle weakness on one side of the body \cite{duncan2005management}. Hemiparesis leads to an asymmetric gait, which negatively impacts an individual's ability to perform activities of daily living by increasing energy expenditure and lowering overall activity levels \cite{jimenez_admittance_2019}\cite{Cho_one_arm}.

To compensate for abnormal gait patterns, most stroke survivors rely on walking aids, with around $76$\% of them using at least one aid after rehabilitation \cite{gosman2002use}. These aids, such as walkers, canes, and crutches, help increase postural stability and reduce weight load on the more affected side, thereby decreasing the risk of falls \cite{Cho_one_arm}.

Conventional walkers are widely used to support post-stroke patients during rehabilitation. While they offer partial body weight support and stability, they have significant limitations; they do not fully prevent falls, may not be safe for severely impaired users, and can become challenging to handle when cognitive impairments are present \cite{Neto2015}.
Additionally, they often increase the user's energy expenditure \cite{sierra_m_humanrobotenvironment_2019}, and lack advanced features like navigation support.

To address these shortcomings, integrating robotics into conventional walkers has led to the development of Smart Walkers (SWs). These robotic devices go beyond basic mobility aid by providing sensory feedback, intelligent navigation assistance, health monitoring, and feedback to caregivers, offering a more comprehensive solution for enhancing safety and rehabilitation outcomes for users with severe physical or cognitive impairments \cite{jimenez_admittance_2019,sierra_m_humanrobotenvironment_2019,Javad_Mehr,DING2022102821,tao2020modeling}.

While SWs address several limitations of conventional walking aids, they often still rely on the user's ability to use both hands effectively. In Jimenez \textit{et al.} \cite{jimenez_admittance_2019}, the authors have employed an admittance controller that calculates forward velocity by summing the forces applied to both handles. The difference in these forces estimates the user-exerted torque, which is used to determine the walker's turning speed. In McLachlan \textit{et al.}\cite{1501134}, the UTS assistant (University of Technology, Sydney assistant) uses strain gauges on handlebars to measure user intent for travel speed and turn rate. Summing the signals determines linear motion, while the difference indicates turning rate. These signals are converted to a dynamic goal position in Cartesian coordinates.

For many stroke survivors with hemiparesis, however, sensory and motor deficits in one upper limb, particularly in the distal regions\cite{Sim2015ImmediateEO}, limit their ability to operate these devices as intended. Addressing these combined challenges of gait and upper limb impairment is crucial to improving mobility and independence in stroke survivors\cite{Cho_one_arm}. 

In recent years, a few approaches have been developed to overcome these challenges, each aiming to address different aspects of user support and control. In Cho \textit{et al.} \cite{Cho_one_arm}, a one-arm motorized walker has been developed to assist hemiplegic stroke survivors by allowing them to drive and steer using buttons on a single handle, reducing the physical effort needed for movement. While this design offers motorized support, it relies on continuous grip and fine motor skills, which can be difficult for users with limited dexterity or fatigue. Additionally, the button-based controls introduce cognitive load and are not intuitive for all users. The single-handle design also prevents the use of the other hand for additional support, even if it is partially functional. 

The in-Hand Admittance Controller (i-HAC) \cite{itadera_-hand_2022} enables control of a robotic walker using tactile grasping with one-handed steering and two-handed rotation through a custom skin sensor system. While it reduces effort, i-HAC relies on precise sensor feedback and an ideal condition, where forces must be applied normally to the sensors. Additionally, the requirement for bilateral control for rotation limits its usability for users with unilateral impairments.

\begin{figure}[!t]
    \centering
    \includegraphics[width=\linewidth]{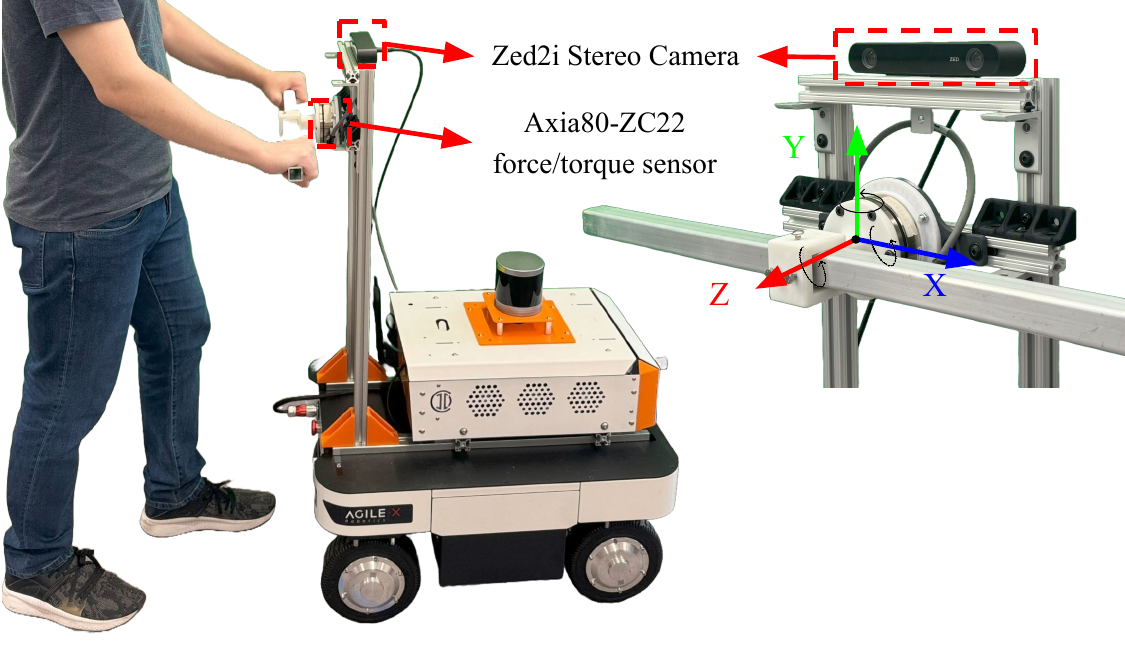}
    \caption{(\textit{left}) Illustration of mobility assistance with the Smart Walker - (\textit{right}) Coordinate reference frame on the force/torque sensor}
    \label{fig:setup}
\vspace{-15pt}
\end{figure}

To overcome the reliance on bilateral upper limb functionality in existing smart walkers, we propose a novel fuzzy control algorithm that utilizes shoulder abduction angles as intuitive indicators of user intention. Fuzzy control is particularly well-suited for this application due to its ability to handle variability and imprecision in user inputs, which are common in individuals with hemiparesis, as noted in Meyers \textit{et al.}\cite{MEYERS2022e14}. This flexibility allows the system to adapt to different levels of user ability and fatigue over time, ensuring consistent and reliable performance. Additionally, fuzzy logic facilitates smooth transitions between control states, enabling the walker to respond seamlessly to subtle shoulder movements. This results in more precise and natural control, enabling users to effectively control the walker using only one functional hand. 

Our system integrates real-time data from a force sensor and camera to interpret shoulder movements of the unaffected arm, assisting with turning and forward motion. Although the walker has a conventional design with two handlebars, this study specifically addresses scenarios where users rely solely on their non-paretic arm due to unilateral upper limb impairment. While both arms could enhance stability, our focus is on improving control and usability for those limited to using one functional arm. This method reduces cognitive load and physical strain, minimizing fatigue and the risk of musculoskeletal injury.

The system has been tested in simulated environments and compared to a traditional torque-based admittance controller. Results indicate that our fuzzy control approach reduces control inaccuracies and user effort by enabling smoother turns and lowering wrist torque and cognitive load. These findings promise for more adaptive and personalized smart walker solutions.

% Change this in future
The rest of the paper is organized as follows: Section \ref{sec2} describes the studied system, computer vision approach for detecting body key-points, the wheeled mobile robot used as a Smart Walker, the proposed controller, the data collection and analysis process. Section \ref{sec3} presents the experimental results with the Smart Walker and discusses the findings. Finally, Section \ref{sec4} discusses outcome of the research, limitations of the study, and directions for future work.

\begin{table*}[!t]
\centering
\caption{Shoulder Abduction Angles in Degrees for Three Directions of Movement ($\mu \pm \sigma$: Mean $\pm$ Standard Deviation) and t-test Results}
\label{tab:angles}
\resizebox{0.9\textwidth}{!}{
\begin{tabular}{c ccc ccc}
\toprule
\multirow{2}{*}{User} & \multicolumn{3}{c}{Direction of Movement} & \multicolumn{3}{c}{t-Test Results} \\ \cmidrule(lr){2-4} \cmidrule(lr){5-7}
 & $L$ & $S$ & $R$ & L vs. S & S vs. R & \\ \cmidrule(lr){2-4} \cmidrule(lr){5-7}
 & $\mu \pm \sigma$ & $\mu \pm \sigma$ & $\mu \pm \sigma$ & $p$-value, $t$-stat & $p$-value, $t$-stat \\
\midrule\midrule
$1$ & $25.97 \pm 5.13$ & $19.64 \pm 1.89$ & $15.04 \pm 2.78$ & $p < 0.05, t = 02.44$ & $p < 0.05, t = 35.09$ \\
$2$ & $25.59 \pm 8.49$ & $20.48 \pm 1.19$ & $17.81 \pm 2.23$ & $p < 0.05, t = 18.43$ & $p < 0.05, t = 42.59$ \\
$3$ & $27.97 \pm 8.42$ & $18.69 \pm 1.80$ & $16.01 \pm 4.26$ & $p < 0.05, t = 14.53$ & $p < 0.05, t = 60.22$ \\
$4$ & $32.27 \pm 7.56$ & $24.02 \pm 4.67$ & $22.92 \pm 3.51$ & $p < 0.05, t = 03.40$ & $p < 0.05, t = 49.44$ \\
$5$ & $39.06 \pm 1.57$ & $27.17 \pm 0.66$ & $20.34 \pm 2.88$ & $p < 0.05, t = 02.85$ & $p < 0.05, t = 106.4$ \\
\bottomrule
\end{tabular}}
% \vspace{-15pt}
\end{table*}

\section{Methodology}
\label{sec2}

In this section, we provide an overview of our smart walker system, detailing its hardware components and the implementation of a conventional admittance controller. We then discuss the limitations of this controller for individuals with unilateral upper limb impairments, highlighting challenges in one-handed operation. To address these challenges, we have analyzed the shoulder abduction angle and demonstrated its strong correlation with user rotational intention. This analysis forms the basis for our proposed vision-based fuzzy control algorithm, designed to enhance usability for users with unilateral impairments.

\subsection{System overview}

Our experimental setup utilizes a Ranger Mini V$3$ mobile base (AgileX Robotics, China) equipped with a Double Ackermann steering system — a configuration used in automotive steering that allows for smooth turning. We modified this mobile base into a smart walker by adding a custom-designed support structure with adjustable handles, allowing for height adaptation to accommodate different users. An Axia$80$-ZC$22$ $6$-DOF force/torque sensor (ATI Industrial Automation, USA) is installed on the handles to measure user interaction forces. Additionally, a Zed$2$i stereo camera (Stereolabs, USA) is mounted on the support structure, facing the user's upper body for real-time pose detection and intention estimation. The camera is calibrated to the specific setup and environment to ensure accurate tracking and minimize errors in pose estimation. Fig. \ref{fig:setup} illustrates the smart walker setup.

\subsection{Conventional Controller}

To establish a baseline for comparison, we implemented a conventional admittance controller based on the work of Sierra et al. \cite{sierra_m_humanrobotenvironment_2019}. This controller converts user-applied forces and torques into linear and angular velocities of the smart walker:
\begin{equation}
L(s) = \frac{v(s)}{F(s)} = \frac{\frac{1}{m}}{s^2 + \frac{b_l}{m}s + \frac{k_l}{m}}
\label{eq:Lin_adm}
\end{equation}

\begin{equation}
A(s) = \frac{\omega(s)}{\tau(s)} = \frac{\frac{1}{J}}{s^2 + \frac{b_a}{J}s + \frac{k_a}{J}}
\label{eq:Ang_adm}
\end{equation}
It uses mass-damper-spring model to create a natural interaction, with the linear system defined by virtual mass ($m$), damping ($b_l$), and elasticity ($k_l$), and the angular system by virtual inertia ($J$), angular damping ($b_a$), and elasticity ($k_a$). Adjusting these parameters tunes the walker’s response.

However, these torque-based approaches are unsuitable for users with only one functional arm. When a user pushes the handlebar forward with one hand to move straight, an unintended torque is generated, causing the walker to rotate undesirably. This requires the user to exert additional wrist torque to counteract the rotation, leading to increased physical strain and potential discomfort. Preliminary experiments with participants simulating unilateral impairment showed an average increase of $81.1$\% in wrist torque compared to bilateral operation when turning right. Further details are provided in the Experiments and Results section.

To address these limitations, there is a need for an alternative control method that does not solely rely on bilateral upper limb input. To address this, we explored vision-based rotation intention detection, which leverages the user's shoulder movements to control the walker.

\subsection{Vision-Based Rotation Intention Detection}

To improve the control of smart walkers for stroke survivors with unilateral impairments, it is critical to accurately detect a user’s walking intention, particularly rotation. Existing methods for estimating walking intention include wearable sensor-based and force/torque-based approaches, each with specific advantages and limitations.

Wearable sensors, such as IMUs, EMGs, or pressure sensors, can precisely monitor joint angles, muscle activity, and motion dynamics. However, they need to be mounted on the user's body and dismounted after use, which is time-consuming. Additionally, they often cause discomfort, require frequent calibration, and may not be reliable for users with unilateral impairments, such as hemiparesis, due to inconsistent data from the affected side\cite{FITTS1989225}.

Force-based systems detect user input forces through sensors embedded in assistive devices, such as the handles of a walker, to infer movement intentions. While these systems are simple and provide real-time feedback, they can be problematic for users with one functional limb, as uneven force distribution can confuse the controller and increase strain on the functional hand \cite{chalaki2024evaluating}.

However, vision-based approaches offer a non-invasive alternative by using cameras and advanced algorithms to detect body posture and subtle movements that indicate a user's walking intention. Unlike wearable or force-based methods, vision-based systems do not require physical contact or extensive calibration and can provide detailed, real-time information about a user’s body posture\cite{POPPE20074,soleymani2021surgical,soleymani2024hands}.

\begin{table}[t!]
\centering
\caption{Averages of Pearson Correlation Coefficient\\($SR$: Straight $\rightarrow$ Right Turn, $SL$: Straight $\rightarrow$ Left Turn)}
\resizebox{0.43\textwidth}{!}{
\begin{tabular}{llccccc}
\toprule
 &  & \multicolumn{5}{c}{User} \\ 
\cmidrule(lr){3-7} 
 &  & 1 & 2 & 3 & 4 & 5 \\ 
\midrule\midrule
\multirow{2}{*}{Task} & $SL$  & 0.87 & 0.84 & 0.92 & 0.89 & 0.95 \\ 
                      & $SR$ & 0.86 & 0.69 & 0.76 & 0.88 & 0.91 \\ 
\bottomrule
\end{tabular}}
\vspace{-15pt}
\label{tab:pearson}
\end{table}

In this study, we propose a vision-based method using the Zed$2$i stereo camera to detect user walking intentions for a smart walker. This system tracks the $3$D positions of key body landmarks in real-time, such as the shoulder, elbow, and hip joints. To account for stroke survivors with hemiparesis, who may have limited use of one upper limb, our approach focuses on tracking only the intact side of the body. For consistency and clarity in this research, we assume the non-paretic side is the right side of the body. By monitoring these points on the right side, we calculate the shoulder abduction angle, which is shown later to have a strong correlation with the user’s intention to turn or move straight.

\begin{figure*}[!t]
    \centering
    \includegraphics[width=\linewidth]{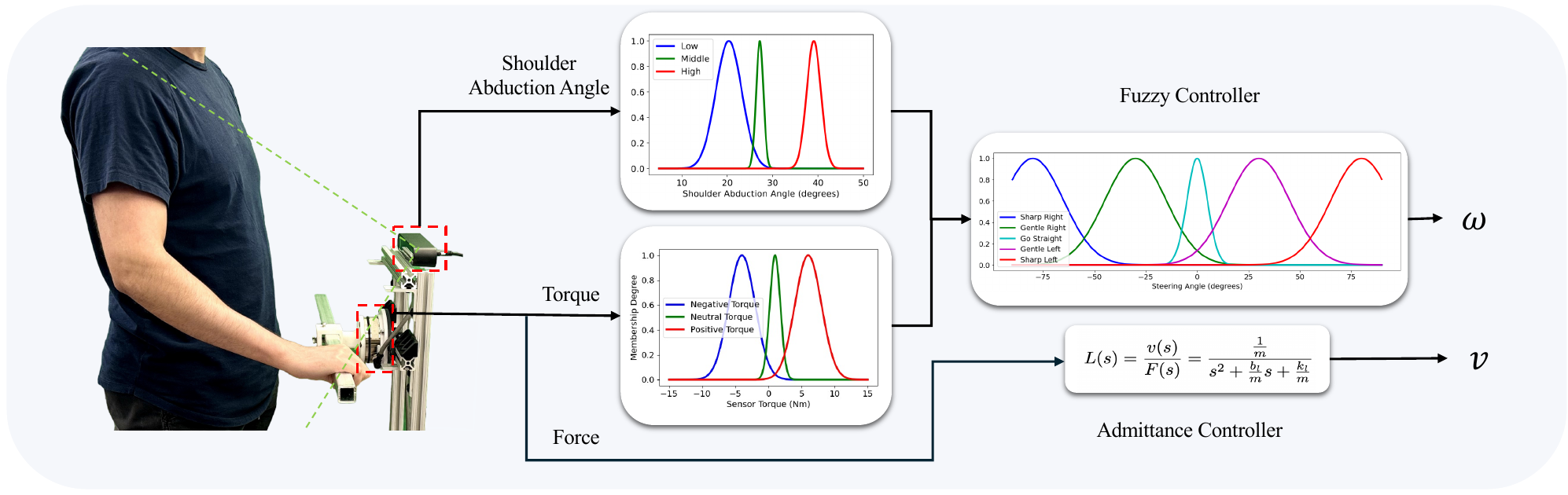}
    \caption{Block Diagram Illustrating Key Components of the Smart Walker Control Framework}
    \label{fig:Block}
\vspace{-10pt}
\end{figure*}

% The body tracking algorithm first detects 2D joint positions using a convolutional neural network (CNN) on the RGB image stream provided by the Zed2i.
The body tracking algorithm utilizes the StereoLabs Positional Tracking API, specifically employing the \texttt{HUMAN\_BODY\_ACCURATE} model, which offers the highest accuracy among available models \cite{stereolabs2023}. This model is applied to the RGB image stream provided by the Zed2i camera. For representing the human body, the \texttt{BODY\_18} format is used, which includes $18$ keypoints following the COCO18 skeleton representation \cite{COCO}. Then, leveraging the stereo depth map, the $2$D keypoints are mapped to $3$D space. Specifically, for each detected keypoint, the corresponding depth value \( Z \) is retrieved, and the $2$D pixel coordinates \( (x, y) \) are transformed into $3$D coordinates \( (X, Y, Z) \) using the camera’s intrinsic parameters: 
\( X = ((x - c_x) \cdot Z) / f_x \) and \( Y = ((y - c_y) \cdot Z) / f_y \), where \( c_x \) and \( c_y \) are the principal point offsets (the optical center), and \( f_x \) and \( f_y \) are the focal lengths of the camera in pixels.

Using the obtained $3$D keypoints, we compute angles between joints to infer body posture and user intention. Specifically, the shoulder abduction angle is determined by examining the relationship between two vectors: one extending from the right shoulder to the right elbow, and the other from the right shoulder to the right hip. This calculation, based on the cosine rule, assesses the orientation of these vectors relative to one another. By analyzing changes in this angle, we can determine whether the user intends to turn or move straight

A frame rate of approximately $50$ fps ensures that motion is tracked smoothly, while its depth-sensing capabilities allow us to accurately detect body posture even in dynamic environments. The real-time body tracking data is fed into a fuzzy controller, which interprets the user's motion intention and adjusts the smart walker’s angular speed accordingly. This system significantly reduces the cognitive load and physical strain required to operate the walker, enabling users to control it with minimal effort from a single hand.

\subsection{Fuzzy Control}

Developing a control system for a smart walker that accurately detects user intentions (especially for individuals with unilateral upper limb impairments) is challenging with traditional controllers. These controllers often rely on direct force or torque inputs, which users may struggle to provide consistently, leading to unreliable intention detection. Fuzzy logic control offers several advantages in this context: it is robust to sensor noise and variability, allows for intuitive modelling based on human reasoning, and provides smooth, adaptive control actions. Leveraging these benefits, we implement a fuzzy control algorithm that utilizes two primary inputs to determine the smart walker's turning rate, while maintaining linear velocity generation through a conventional admittance controller. This hybrid approach enhances both responsiveness and intuitiveness for users with unilateral impairments.

The fuzzy controller utilizes two primary inputs:
\begin{itemize}

\item \textbf{Exerted Torque on the Handles}: Defined by the linguistic terms \textit{Negative}, \textit{Neutral}, and \textit{Positive}, this input represents the overall torque applied by the user on the walker handles.

\item \textbf{Shoulder Abduction Angle}: Represented by the linguistic terms \textit{Low}, \textit{Middle}, and \textit{High}, this input indicates the angle of abduction of the user's shoulder, which correlates with their intention of turning.
\end{itemize}
The fuzzy membership functions for both inputs are constructed using Gaussian shapes, which provide smoother transitions between fuzzy sets and offer better handling of noise and small input variations compared to the sharper transitions seen with trapezoidal and triangular functions.

The controller output is the angular speed command, described by linguistic terms such as \textit{Sharp Left}, \textit{Gentle Left}, \textit{Go Straight}, \textit{Gentle Right}, and \textit{Sharp Right}, covering a range of $-90$ to $90$ degrees per second, which corresponds to the maximum steering capabilities of the walker.

\begin{table}[!t]
\centering
\caption{Fuzzy Control Rules for Steering Based on Right Shoulder Abduction Angle and Sensor Torque}
\label{tab:Rules}
\resizebox{0.48\textwidth}{!}{
\begin{tabular}{cccc}
    \toprule
    \multirow{2}{*}{\shortstack{Right shoulder \\ abduction angle}} & \multicolumn{3}{c}{Torque $Y$} \\
    \cmidrule(lr){2-4}
    & Negative & Neutral & Positive \\
    \midrule\midrule
    Low & \textcolor{red}{Sharp} Right & \textcolor{black}{Straight} & \textcolor{black}{Straight} \\
    Middle & \textcolor{orange}{Gentle} Right & \textcolor{black}{Straight} & \textcolor{black}{Straight} \\
    High & \textcolor{orange}{Gentle} Right & \textcolor{orange}{Gentle} Left & \textcolor{red}{Sharp} Left \\
    \bottomrule
\end{tabular}}
\vspace{-15pt}
\end{table}

\begin{table*}[!t]
\centering
\caption{Performance comparison between Conventional Controller ($CC$) and Proposed Controller ($PC$) for left ($L$), right ($R$), and straight ($S$) movement directions: Exerted torque by the right wrist in N.m, presented as $\mu\pm\sigma$ (Mean $\pm$ $std$).}
\label{tab:torque}
\resizebox{0.85\textwidth}{!}{
\begin{tabular}{ccccccc}
\toprule
\multirow{2}{*}{User}&\multicolumn{6}{c}{Direction of Movement}\\
\cmidrule(lr){2-7}
&\multicolumn{1}{c}{$L-CC$}&\multicolumn{1}{c|}{$L-PC$}&\multicolumn{1}{c}{$S-CC$}&\multicolumn{1}{c|}{$S-PC$}&\multicolumn{1}{c}{$R-CC$}&\multicolumn{1}{c}{$R-PC$}\\
\midrule\midrule
1    & $-2.78 \pm 1.13$  & $-0.42 \pm 0.54$  & $-4.02 \pm 0.91$  & $-0.82 \pm 0.13$  & $-6.89 \pm 0.52$  & $-1.36 \pm 0.33$  \\ 
2    & $-0.11 \pm 2.00$  & $-0.31 \pm 0.36$  & $-4.76 \pm 0.46$  & $-0.54 \pm 0.16$  & $-5.58 \pm 0.87$  & $-0.76 \pm 0.35$  \\ 
3    & $-2.92 \pm 0.95$  & $-0.55 \pm 0.41$  & $-3.85 \pm 0.54$  & $-0.72 \pm 0.08$  & $-6.31 \pm 0.95$  & $-1.22 \pm 0.29$  \\ 
4    & $-0.91 \pm 1.03$  & $-0.65 \pm 0.42$  & $-5.41 \pm 0.43$  & $-0.78 \pm 0.19$  & $-7.51 \pm 0.44$  & $-1.44 \pm 0.28$  \\ 
5    & $-1.03 \pm 0.54$  & $-0.51 \pm 0.39$  & $-2.13 \pm 0.10$  & $-0.71 \pm 0.26$  & $-5.52 \pm 1.48$  & $-1.23 \pm 0.25$  \\

\bottomrule
\end{tabular}}
% \vspace{-15pt}
\label{torque_comp}
\end{table*}

Operating based on a set of fuzzy rules (see Table \ref{tab:Rules}), the controller maps combinations of input linguistic terms to output steering commands, enabling gradual and smooth adjustments. For example, if a user exhibits a high shoulder abduction angle (\textit{High}) and applies negative torque (\textit{Negative}), the controller interprets this as an intention to make a \textit{Sharp Right} turn. This fuzzy logic framework allows for easy modification of the rule base for rapid adjustments and user configurability to cater to individual needs, enhancing the system's adaptability.

By providing a flexible and intuitive control interface, the fuzzy controller significantly improves the usability and effectiveness of the smart walker for users with various impairments, addressing the limitations of traditional control methods.

\subsection{Data Collection}

The study involved five healthy volunteers with varying ages ($23$ to $35$ years), heights ($168$ to $185$ cm), and weights ($61$ to $86$ kg) to achieve more generalized results, approved by the Research Ethics Boards of the University of Alberta under reference number Pro$00139670$.

The experiments were carried out in a controlled laboratory and the smart walker was adjusted to each participant's height for comfort. Before each session, the force/torque sensor was calibrated to ensure accurate torque measurements and the Zed2i camera was mounted at a height of $1.1$ meter to optimally capture shoulder movements.

Participants were asked to perform a series of tasks using only one hand on the walker, including walking straight for $5$ meters and performing $90$-degree left or right turns. Each task was repeated five times to ensure data reliability. Participants were given rest periods between tasks to prevent fatigue. During the experiments, the abduction angles of the operating hand shoulder and the force and torque exerted on the handle were recorded at $50$ Hz. To focus exclusively on steady-state gait, the initial acceleration and final deceleration phases were excluded from the analysis.

\subsection{Data Analysis}

During each walking trial, torque on the walker handles and shoulder abduction angles were smoothed using a moving-average filter with a window size of $60$ to minimize gait-induced cyclical effects \cite{MOSTAFAVI2022102173}. A Pearson correlation analysis was conducted to examine the relationship between shoulder abduction angles and handle torque, verifying if these movements reliably predict turning intentions.

Data were segmented into walking straight, executing a $90$-degree turn, and resuming straight walking, with averages calculated for each segment to determine typical shoulder abduction angles for various movements. These findings helped set thresholds for the fuzzy controller and were used in paired t-tests to identify significant differences in shoulder angles across tasks.

\section{EXPERIMENTS AND RESULTS}
\label{sec3}

\subsection{Experiment 1: Correlation Analysis of Shoulder Abduction Angles and Rotational Intentions}

The first experiment aimed to assess the correlation between shoulder abduction angle and the torque exerted by users on the walker's handle. Each participant was instructed to operate the walker with one hand and complete two specific tasks five times. The first task involved following a path marked by two lines on the flat ground, consisting of a $4$-meter straight segment, a $90$-degree right turn, followed by another $4$-meter straight segment. The second task required participants to navigate the same path from the opposite end, resulting in a straight segment, a $90$-degree left turn, and another straight segment. The conventional control system was used for this experiment to evaluate how naturally users could manipulate the walker’s movement through shoulder and arm dynamics.

In this experiment, we analyzed the torque measured by the sensor in the z-direction (refer to Fig. \ref{fig:setup}) as an indicator of the user's intent to turn, along with the shoulder abduction angle. The Pearson Correlation test, detailed in Table \ref{tab:pearson}, revealed a high correlation across different users' data, affirming that these variables linearly relate. This correlation supports the selection of these measures for our fuzzy control system, suggesting that shoulder movements can effectively predict turning intentions.

\begin{table}[!t]
\centering
\caption{Average Likert Scale Ratings for Conventional vs. Fuzzy Controllers}
\label{tab:questionnaire_results}
\begin{tabular}{lcc}
\toprule
Dimension & Conventional Controller & Fuzzy Controller \\ 
\midrule\midrule
Usability & $1.46 \pm 0.64$ & $4.20 \pm 0.56$ \\ 
Comfort & $1.00 \pm 0.00$ & $4.00 \pm 0.71$ \\ 
Cognitive Load & $1.13 \pm 0.41$ & $4.07 \pm 0.70$ \\ 
Physical Strain & $1.07 \pm 0.26$ & $4.40 \pm 0.72$ \\ 
Overall Satisfaction & $2.30 \pm 1.42$ & $4.10 \pm 0.74$ \\ 
Preference ($\%$) & $0\%$ prefer Conventional & $100\%$ prefer Fuzzy \\ 
\bottomrule
\end{tabular}
\vspace{-15pt}
\label{Likert_table}
\end{table}

To validate the differences in shoulder abduction angles for straight walking and turning (left and right), paired t-tests were applied. The results, shown in Table \ref{tab:angles}, demonstrated significant differences in angles between these tasks, confirming that shoulder abduction angles vary distinctly with the direction of movement. This sensitivity to directional intent is crucial for calibrating the fuzzy control system effectively.

Based on the findings for each individual participant, taking into account differences in body size parameters, we set personalized thresholds for the fuzzy control system’s membership functions, based on what is shown in Table \ref{tab:angles}.
Specifically for user $5$, the average shoulder angles were set at $27.17$ degrees for straight walking, $39.06$ degrees for turning left, and $20.34$ degrees for turning right.
For torque inputs, thresholds were established at $1$ Nm for straight walking, $6$ Nm for turning left, and $-4$ Nm for turning right.
These settings allow the fuzzy controller to accurately interpret user inputs and adjust the walker's movements, enhancing its responsiveness and usability.

\subsection{Experiment 2: Quantitative and Qualitative Analysis of the Proposed Fuzzy Controller}
This experiment evaluated the effectiveness of the proposed fuzzy control system, as detailed in the Fuzzy Control subsection, calibrated using the thresholds established in the previous experiment. The same five participants were asked to complete the same tasks as before, but this time using the smart walker equipped with the fuzzy controller.

To assess improvements quantitatively, we compared the average wrist torque generated by each user during movement in different directions between the conventional and the new fuzzy controller (see Table \ref{torque_comp}). The results revealed a significant reduction in wrist torque—$55.89$\% for turning left, $80.36$\% for going straight, and $81.17$\% for turning right—indicating less user effort, which would lead to reduced fatigue and a lower risk of wrist injury when controlling the robot.

\begin{figure}[t!] \begin{center} \resizebox{1\hsize}{!}{\includegraphics*{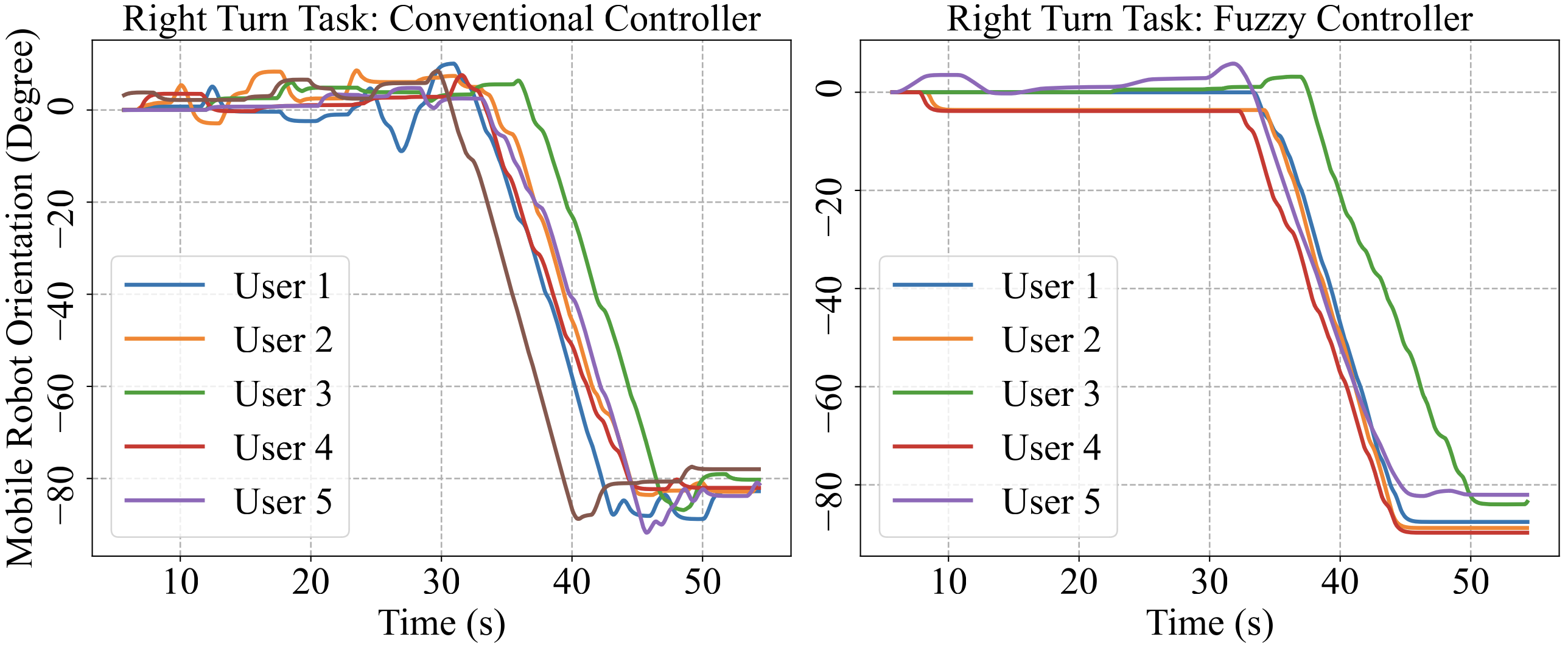}} \caption{Mobile robot orientation under conventional and proposed vision-based fuzzy controller, steering with one hand.} \label{interv-conn} \end{center} \vspace{-20pt} \end{figure}

In addition to the torque analysis, the orientation control of the smart walker was noticeably smoother with the fuzzy controller. The proposed system maintained almost stable  orientation with less fluctuations, responding fluidly to user input. In contrast, the conventional controller exhibited frequent changes in orientation (see Fig. \ref{interv-conn} for turning right results). This difference, particularly when turning right as the user's wrist is bearing larger torque load, is a paramount improvement.

To further evaluate user experience, a $5$-point Likert scale questionnaire was used to assess six aspects: usability, comfort, cognitive load, physical strain, overall satisfaction, and preference. The results, shown in Table \ref{Likert_table}, indicate that the fuzzy controller outperformed the conventional system in all aspects. Participants described the fuzzy controller as more intuitive and easier to use, highlighting reduced effort required for maneuvering.

Together, these findings—quantitative data and user feedback—demonstrate that the fuzzy control system enhances overall usability and reduces both cognitive and physical demands, aligning with our goal of developing a more intuitive and supportive assistive device for individuals with unilateral impairments.

\section{CONCLUSIONS AND FUTURE WORK}
\label{sec4}

This study introduced a novel approach to enhancing Smart Walkers (SWs) control systems for stroke survivors with unilateral impairments. By incorporating a fuzzy control algorithm that utilizes shoulder abduction angles, our system effectively interprets user intentions using only one functional hand. The integration of real-time data from force sensors and a stereo camera facilitated intuitive and responsive control, significantly reducing the cognitive load and physical strain associated with traditional walker operation.

Experimental results demonstrated the efficacy of our proposed fuzzy controller, revealing significant improvements in user comfort and a reduction in wrist torque compared to conventional admittance controllers. These findings validate the potential of our system to improve mobility and independence for individuals with hemiparesis and affirm shoulder abduction angles as reliable indicators of turning intentions.

However, the current system has limitations that need to be addressed. One major limitation is the reliance on visual systems, which can be unreliable under varying lighting conditions, affecting the accuracy of shoulder angle measurements. Additionally, the fuzzy controller uses fixed thresholds that may not be optimal for all users. Developing adaptive methods to adjust these thresholds based on individual user characteristics and preferences is needed to enhance system performance.

Looking ahead, we plan to enhance the functionality of smart walkers by exploring Deep Learning and Reinforcement Learning algorithms, which would allow the walker to adapt to individual user behaviors and preferences over time, providing more personalized assistance \cite{ou2024autonomous,soleymani2021deep,chalaki2021deep}. Additionally, we plan to conduct larger-scale studies with a broader population of stroke survivors, particularly those with unilateral upper limb impairments, to validate and refine the system. Our ultimate goal is to develop a more adaptable and personalized smart walker that enhances mobility, independence, and quality of life for individuals with mobility impairments.


\bibliographystyle{ieeetr}
\bibliography{Papers}

\end{document}